\documentclass{dukecei}

\usepackage{helvet}
\usepackage{courier}

\usepackage{amsmath,amsfonts,bm}

\def\eqref#1{equation~\ref{#1}}

\def\1{\bm{1}}

\DeclareMathAlphabet{\mathsfit}{\encodingdefault}{\sfdefault}{m}{sl}
\SetMathAlphabet{\mathsfit}{bold}{\encodingdefault}{\sfdefault}{bx}{n}

\usepackage{threeparttable}
\usepackage{microtype}
\usepackage{epsfig}

\usepackage[table,xcdraw,dvipsnames]{xcolor}

\usepackage{float}
\usepackage{stfloats}
\usepackage{enumitem}
\usepackage{xstring}
\usepackage{xspace}
\usepackage{makecell}

\usepackage[hang,flushmargin]{footmisc}

\usepackage{mdframed}
\usepackage{tikz}
\usepackage{etoolbox}
\usepackage[capitalize,noabbrev]{cleveref}
\usepackage{thmtools}

\newif\ifreview

\newif\ifarxiv
\newcommand{\arxiv}{\arxivtrue}

\newif\ifcamera

\newif\ifrebuttal

\arxiv

\ifcamera
  \usepackage[accsupp]{axessibility}
\fi

\ifarxiv
  
\fi

\definecolor{OurHL}{RGB}{242,242,242}
\definecolor{OurCrim}{HTML}{801C25}

\definecolor{accentpink}{HTML}{F00080}

\newcommand{\redhead}[1]{%
  \textcolor{accentpink}{\textbf{\textsc{#1}}}%
}

\surroundwithmdframed[
  linewidth=0.5pt,
  linecolor=gray!20,
  tikzsetting={dash pattern=on 3pt off 3pt},
  backgroundcolor=gray!10,
  roundcorner=4pt,
  innerleftmargin=6pt,
  innerrightmargin=6pt,
  innertopmargin=8pt,
  innerbottommargin=6pt
]{theorem}

\surroundwithmdframed[
  linewidth=0.5pt,
  linecolor=gray!20,
  tikzsetting={dash pattern=on 3pt off 3pt},
  backgroundcolor=gray!10,
  roundcorner=4pt,
  innerleftmargin=6pt,
  innerrightmargin=6pt,
  innertopmargin=8pt,
  innerbottommargin=6pt
]{proposition}

\title{
Joint Branch-Space Transform Coding for Diffusion Activation
Quantization with Classifier-Free Guidance
}

\author{%
Mingrun Jiang$^{1,*}$,
Yuejia Liu$^{1,*}$,
Zishan Shao$^{1,*\dagger}$,
Ting Jiang$^{2}$,
Qinsi Wang$^{1}$,
Hancheng Ye$^{1}$,
Yixiao Wang$^{1}$,\\
Rui-Feng Wang$^{3}$,
Kangning Cui$^{4}$,
Yixuan Chen$^{5}$,
Fan Yang$^{4}$,
Xiang Cheng$^{1}$,
Hai Li$^{1}$,
Yiran Chen$^{1}$
\\[0.65em]
\normalfont\small
$^{1}$Duke University
\quad
$^{2}$Carnegie Mellon University
\\
$^{3}$University of Florida
\quad
$^{4}$Wake Forest University
\quad
$^{5}$University of Oxford
}

\begin{document}

\maketitle
\thispagestyle{firstpagestyle}

\vspace{-0.8cm}
\begin{figure*}[ht]
    \centering
    \includegraphics[width=0.88\linewidth]{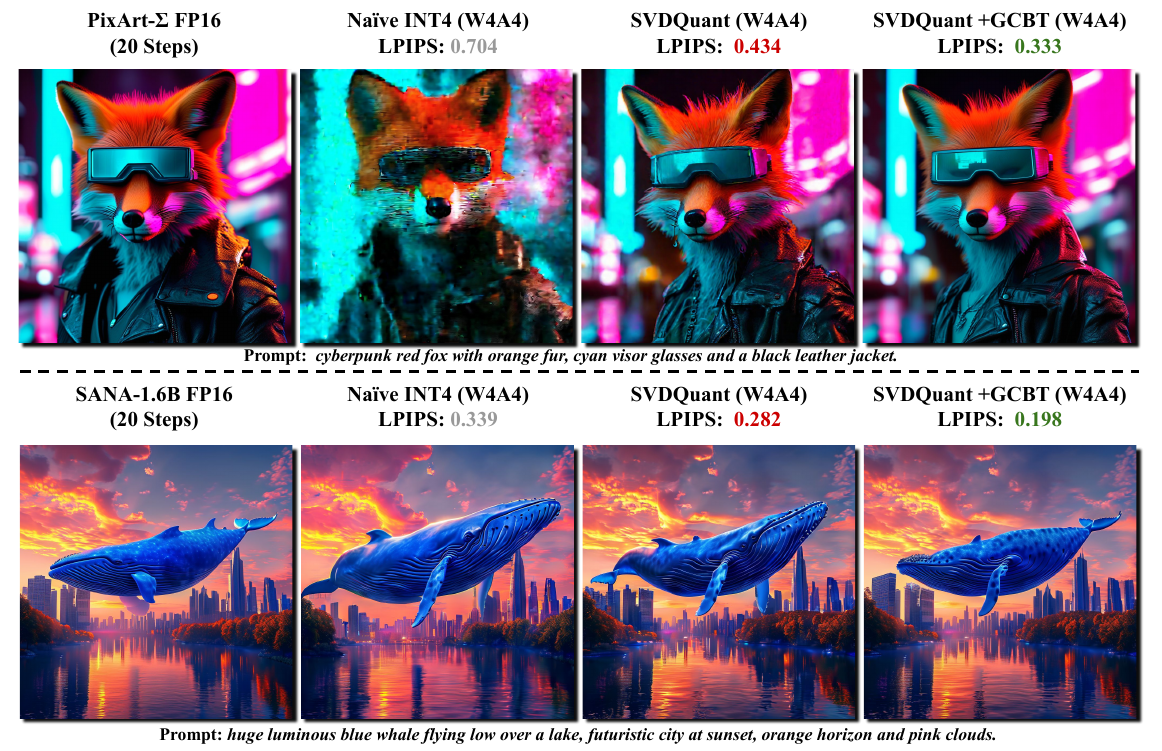}
    \captionof{figure}{
GCBT is a plug-and-play activation transform built on top of existing post-training quantizers in CFG-guided diffusion models. 
By applying a branch-wise rotation prior to quantization, it requires minimal modifications to model parameters or base pipelines. 
On PixArt-$\Sigma$ (top) and Sana-1.6B (bottom), GCBT demonstrates superior visual fidelity over naive PTQ and SVDQuant.
}
    \label{fig:gcbt-comparison}
\end{figure*}

\begin{abstract}
Post-training quantization for diffusion models increasingly exploits timestep, feature, and layer structure. While recent work has begun incorporating CFG structure into diffusion quantization, activation quantization still operates independently across conditional and unconditional coordinates, leaving cross-activation structure unexploited. We show that matched CFG activations form a strongly correlated two-dimensional source and that, under a fixed bit budget, the choice of branch coding basis materially affects quantization fidelity. Motivated by this observation, we introduce \textbf{branch-space transform coding}, which rotates matched CFG branches via an offline derived $2\times2$ orthogonal matrix, requiring minimal modifications to model parameters or the quantization pipeline. We further derive the \textbf{Guidance-Correlation Branch Transform (GCBT)}, which jointly incorporates the CFG guidance direction and cross-branch second moments. Under an equal-rate quantization-noise surrogate, GCBT admits a closed-form per-layer solution without gradient optimization or angle search. Applied on top of existing diffusion PTQ methods, GCBT yields statistically significant fidelity gains in most evaluated comparisons with no statistically significant degradation, while leaving the underlying host quantization pipeline unchanged.
\end{abstract}

\section{Introduction}
\label{sec:intro}

Diffusion models~\citep{sohl2015deep, ho2020denoising, song2020score, nichol2021improved} achieve high generation quality at the cost of repeated, computationally heavy denoising steps. This has motivated extensive work on post-training quantization (PTQ)~\citep{chen2024qdit,li2023qdiffusion,shao2026tr,li2026svdquant,zhao2024viditq,shafi2026closing} to reduce inference overhead via low-bit representations of weights and activations. Recent methods increasingly exploit activation structure during diffusion inference, exploiting (i) variation across denoising timesteps~\citep{chen2024qdit,zhao2024viditq}, (ii) architectural and feature heterogeneity within the denoising model~\citep{li2023qdiffusion,chen2024qdit}, and (iii) activation outliers~\citep{li2026svdquant,li2026deltaquant}. These advances raise a natural question: \textit{what additional structural information in diffusion inference can be exploited to improve activation quantization? }





One such structure is the paired computation in classifier-free guidance (CFG)~\citep{ho2022classifier}.
CFG plays a central role in modern diffusion pipelines by strengthening the influence of the condition on generation.
At each denoising step, CFG evaluates the same model weights with and without condition, producing matched predictions $p_c$ and $p_u$ and then combining them using a guidance scale $s$.
We denote the corresponding activations as $X_c$ and $X_u$, and refer to the two-dimensional axis formed by this matched pair as the \textbf{\emph{CFG branch space}.}




While recent work has begun incorporating CFG into post-training calibration~\citep{shao2026tr,shafi2026closing}, activation quantization still typically operates on $X_c$ and $X_u$ independently in separate branch coordinates. This choice implicitly fixes the branch basis, overlooking the fact that both activations arise from matched CFG passes through the same model weights.




\begin{wrapfigure}{r}{0.38\textwidth}
    \vspace{-12pt}
    \centering
    \includegraphics[width=\linewidth]{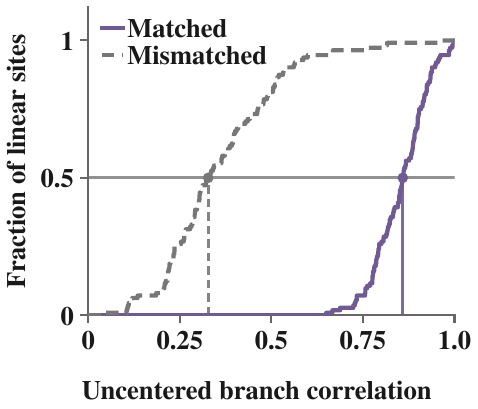}
    \caption{\textbf{Matched CFG branches are strongly correlated.} Matched CFG branches exhibit substantially stronger cross-branch correlation than deliberately mismatched pairs.}
    \label{fig:direction-mismatch}
    \vspace{-10pt}
\end{wrapfigure}



We find two properties of this branch space that are relevant to activation quantization. First, matched $X_c$ and $X_u$ activations are strongly correlated, exhibiting cross-branch correlation coefficients of a median uncentered correlation of 0.86 (vs. 0.33 for mismatched pairs) across DiT-XL/2~\citep{peebles2023scalable} layers (Figure~\ref{fig:direction-mismatch}). Second, the two CFG branches are coupled at the model output by the known guidance direction $g_s = [s, 1-s]^\top$, which varies with the guidance scale $s$. As illustrated in Figure~\ref{fig:branch-space-motivation}, standard PTQ quantizes activations independently along native branch axes, imposing a structural mismatch between the quantization grid, principal activation variance, and the guidance vector $g_s$. Under a fixed activation-bit budget, this directional mismatch limits quantization fidelity.

Motivated by these observations, we introduce \textbf{branch-space transform coding}. Before activation quantization, we apply a $2\times2$ orthogonal branch rotation matrix ($R$) across each matched conditional/unconditional pair, quantize the transformed activations on top of the existing low-bit quantizer, and invert the transform afterward. The procedure does not change model parameters, nominal activation precision, or the number of quantized values; it changes only the branch coordinates in which activation quantization is performed. Branch-space coding is therefore complementary to existing timestep-, feature-, layer-, and precision-aware diffusion PTQ methods.

To determine the optimal rotation matrix $R$, we derive the \textbf{\textit{Guidance-Correlation Branch Transform (GCBT)}}, which jointly incorporates the CFG guidance direction and per-layer cross-branch second moments. Under an equal-rate quantization-noise surrogate, this formulation yields a closed-form per-layer solution without gradient optimization or angle search. Across diverse diffusion PTQ backbones, models, and datasets, GCBT consistently improves reconstruction and perceptual fidelity, establishing CFG branch space as a complementary axis for activation quantization.




Our work makes the following contributions:
\begin{itemize}

    \item \textbf{CFG Branch Space as a Design Axis.}
To our knowledge, we are the first to treat the CFG branch axis as an explicit coding space, demonstrating that cross-branch structure between matched activations directly governs quantization fidelity.

    \item \textbf{Branch-Space Transform Coding \& GCBT.}
    We introduce a lightweight branch-space transform coding framework that incorporates the CFG guidance direction and cross-branch second moments into the coding basis.

    \item \textbf{Closed-Form Global Optimum.}
    We derive a guidance-weighted distortion objective over $2\times2$ orthogonal rotations and prove that its global optimum admits a closed-form per-layer solution under an equal-rate quantization-noise surrogate.

    \item \textbf{Broad PTQ Compatibility.} 
    We apply GCBT on top of existing activation quantizers without modifying the underlying quantization method, achieving consistent improvements in reconstruction and perceptual fidelity.
\end{itemize}

\section{Related Work}
\label{sec:related}

\paragraph{Post-Training Quantization for Diffusion Models.} Diffusion PTQ increasingly exploits structure specific to iterative generation. Q-Diffusion~\citep{li2023qdiffusion} addresses timestep-dependent activation distributions and architectural outliers, while Q-DiT~\citep{chen2024qdit} and ViDiT-Q~\citep{zhao2024viditq} further account for spatial, temporal, layer, and channel variation in Diffusion Transformers. SVDQuant~\citep{li2026svdquant} handles activation outliers through smoothing and low-rank compensation, and DeltaQuant~\citep{li2026deltaquant} exploits spatiotemporal activation structure for low-bit video diffusion. These methods adapt quantization to diffusion-specific activation structure, but retain the native conditional/unconditional branch coordinates during activation quantization.

\paragraph{Rotation-Based Quantization.}
Rotary quantization has been widely studied beyond diffusion models. QuaRot~\citep{ashkboos2024quarot} applies orthogonal rotations to reduce activation outliers, while SpinQuant~\citep{liu2025spinquant} learns rotation matrices for improved low-bit quantization. Related methods use Hadamard-based incoherence processing~\citep{tseng2024quip} or learnable linear transformations~\citep{sun2025flatquant,hu2025ostquant} to improve the quantizability of weights and activations. In diffusion models, TR-DQ~\citep{shao2026tr} introduces timestep-dependent feature rotations, while OrbitQuant~\citep{lee2026orbitquant} quantizes in a normalized rotated basis. These approaches transform feature or hidden-state coordinates; in contrast, our transform acts across the two-dimensional CFG branch axis while leaving token and feature coordinates unchanged.

\paragraph{CFG-Aware Quantization.} Recent diffusion PTQ methods have also begun to exploit the relation between CFG branches. TR-DQ~\citep{shao2026tr} uses redundancy across CFG executions, while guidance-aware methods such as GAMP~\citep{shafi2026closing} calibrate quantization with respect to the guided prediction and CFG sensitivity. These methods incorporate CFG structure while retaining the native conditional/unconditional branch basis. Our work studies a complementary variable: the \emph{branch coding basis} itself. We treat the CFG branch axis as an explicit quantization space and select its basis from the CFG branch direction and cross-branch second moments.

\section{Method}
\label{sec:method}

\subsection{Problem Setup}

Classifier-free guidance (CFG)~\citep{ho2022classifier} combines the
conditional and unconditional branch predictions as
\begin{equation}
p_{\mathrm{cfg}}
=
s p_c + (1-s)p_u
=
g_s^\top
\begin{bmatrix}
p_c\\
p_u
\end{bmatrix},
\qquad
g_s=
\begin{bmatrix}
s\\
1-s
\end{bmatrix},
\label{eq:cfg}
\end{equation}
where $s$ is the guidance scale.

At a given denoising step and activation-quantization site, let
$X_c,X_u\in\mathbb{R}^{N\times d}$ denote the matched conditional and
unconditional branch activations, where $N$ indexes token or spatial
positions and $d$ is the feature dimension. We stack them along the CFG
branch axis as
\begin{equation}
X=
\begin{bmatrix}
X_c\\
X_u
\end{bmatrix}
\in\mathbb{R}^{2\times N\times d},
\label{eq:branch-space}
\end{equation}
whose leading dimension defines the CFG branch space with native coordinates
$(c,u)$. At each token--feature coordinate $(n,j)$, the corresponding
matched branch-activation pair is
$x_{n,j}=[x_{c,n,j},x_{u,n,j}]^\top\in\mathbb{R}^2$.

Let $\mathcal{Q}_A$ denote the host $A$-bit activation quantizer.
Standard PTQ quantizes the two branch activations in the native basis:
$\widehat{X}_{\mathrm{nat}}=
\begin{bmatrix}
\mathcal{Q}_A(X_c)\\
\mathcal{Q}_A(X_u)
\end{bmatrix}$.
\vspace{-2mm}

\begin{figure*}[t]
\centering
\small
\includegraphics[width=0.99\textwidth]{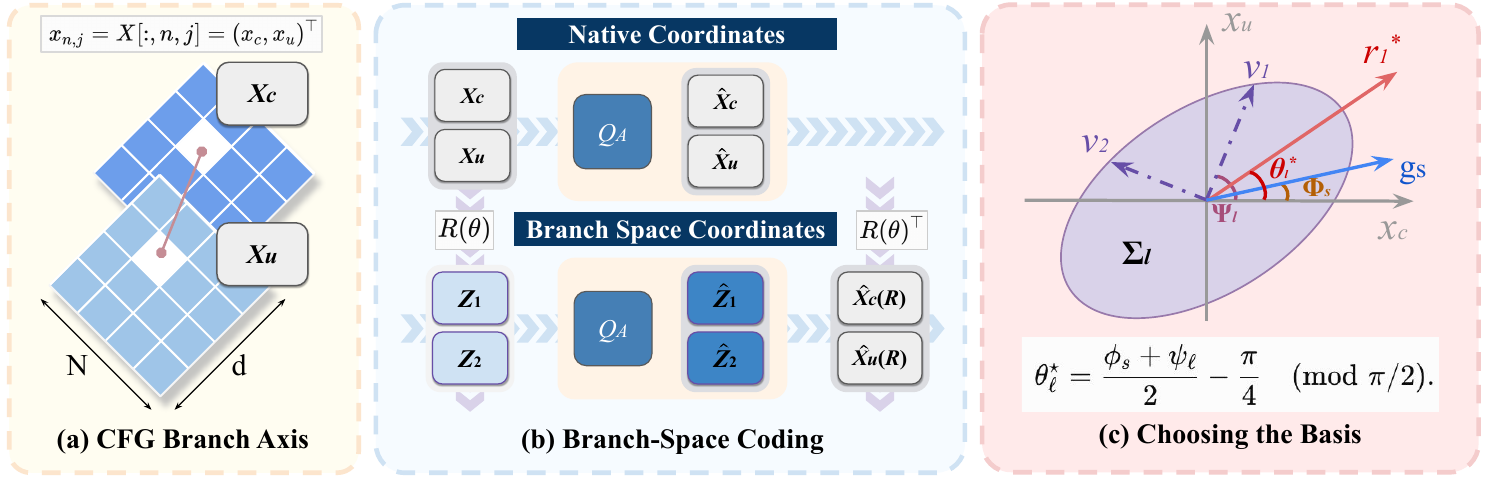}
\caption{
\textbf{Overview of branch-space coding.}
\textbf{(a)} Matched CFG activations define a two-dimensional branch axis at each token--feature coordinate.
\textbf{(b)} Native quantizer $Q_{A}$ operates directly on $(X_c,X_u)$, while branch-space coding rotates the pair, applies the same host quantizer in transformed coordinates, and rotates back.
\textbf{(c)} The coding basis is determined jointly by the activation geometry $\Sigma_\ell$ and CFG branch direction $g_s$, yielding the optimal coding axis $r_1^\star$ and angle $\theta_\ell^\star$ where \(\theta_\ell^\star=\angle(X_c,r_1^\star)\).}
\label{fig:branch-space-motivation}
\end{figure*}

\subsection{Branch-Space Structure and Transform Coding}
\label{sec:branch-space}

Standard diffusion PTQ~\citep{chen2024qdit,li2023qdiffusion,li2026svdquant, li2026deltaquant,lee2026orbitquant,zhao2024viditq} quantizes CFG activations directly in the native
$(c,u)$ branch coordinates. To determine whether activation quantization can benefit from a different branch coding basis, we examine whether matched CFG activations exhibit cross-branch structure and whether changing only the branch coding basis affects quantization fidelity under a fixed activation-bit budget.


\redhead{Observation I: } \textit{\textbf{CFG branches form a structured matched source.}} 
\textit{On DiT-XL/2 with W4A4 SVDQuant (CFG $=2$, 64 trajectories), the median correlation between matched branch activations across all linear sites  in DiT-XL/2 is $0.859$, compared with $0.328$ for deliberately mismatched pairs (Figure~\ref{fig:direction-mismatch}, Table~\ref{tab:motivation-calibration-sweep}). The matched activations are also highly directional: their dominant axis lies at a median $42.71^\circ$ from the conditional axis, close to the diagonal $c=u$. Figure~\ref{fig:branch-space-motivation}(a) illustrates this structure schematically. These measurements motivate considering the matched activations jointly when choosing a quantization basis.}



\redhead{Observation II: } \textit{\textbf{Activation-quantization fidelity depends on the branch basis.}} \textit{Under the same activation-bit budget and host activation quantizer, changing only the branch coding basis produces measurable differences in generation fidelity. At CFG $=4.5$, a guidance-aligned rotation improves PSNR and SSIM while reducing LPIPS under the same host quantizer and activation-bit budget; see Table~\ref{tab:basis-ablation}.}

These observations motivate treating the branch coding basis as an explicit degree of freedom in activation quantization, rather than inheriting the native $(c,u)$ execution coordinates. This leads to the following general formulation. 

\begin{definition}[Branch-space transform coding]
Let $R\in\mathbb{R}^{2\times2}$ be an orthogonal matrix, whose rows define an orthonormal coding basis. Given a host activation
quantizer $\mathcal{Q}_A$, we quantize matched CFG activations in the
transformed branch basis as
\begin{equation}
\widehat{X}(R)
:=
R^\top\mathcal{Q}_A(RX) = \begin{bmatrix}
\widehat{X}_c(R)\\\widehat{X}_u(R)
\end{bmatrix}.
\label{eq:branch-coding}
\end{equation}
\end{definition}
The transform acts only across matched conditional/unconditional pairs,
leaving token and feature coordinates unchanged. Setting $R=I$ recovers
native branchwise activation quantization.

\subsection{Guidance-Correlation Branch Transform}
\label{sec:gcbt}

Branch-space coding leaves one question: \emph{which basis should be used?} We choose the basis jointly from the geometry of matched CFG activations and the CFG branch direction.


At layer $\ell$, for each matched token--feature coordinate $(n,j)$, define
\begin{equation}
x_{n,j}^{(\ell)}
=
\begin{bmatrix}
X_c^{(\ell)}[n,j]\\
X_u^{(\ell)}[n,j]
\end{bmatrix}
\in\mathbb{R}^2.
\label{eq:branch-pair}
\end{equation}
Suppressing $(n,j)$ for brevity, we write a generic matched
branch-activation pair as
$x^{(\ell)}=[x_c^{(\ell)},x_u^{(\ell)}]^\top$.
The branch-space geometry at layer $\ell$ is summarized by
\begin{equation}
\Sigma_\ell
=
\mathbb{E}\!\left[
x^{(\ell)}x^{(\ell)\top}
\right],
\label{eq:sigma}
\end{equation}
where the expectation is taken over calibrated matched branch-activation
pairs. Thus, $\Sigma_\ell$ is the uncentered second-moment matrix describing
the two-dimensional branch geometry at layer $\ell$.



For the remainder of this derivation, we fix layer $\ell$ and suppress the layer superscript on local branch-space quantities. Let $R=[r_1^\top;r_2^\top]\in\mathbb{R}^{2\times2}$ be an orthogonal rotation matrix, where $r_1,r_2\in\mathbb{R}^2$ define the coding basis. For a matched pair $x$, the transformed coordinates are $z=Rx$, with $z_i=r_i^\top x$.

\redhead{Design principle: }\textbf{\textit{Couple activation geometry with guidance sensitivity.}} Let $\hat z_i$ denote the quantized value of transformed coordinate $z_i$. We approximate the quantization distortion of each transformed coordinate as proportional to its projected second moment,
\begin{equation}
\mathbb{E}\!\left[(\hat z_i-z_i)^2\right]
=
\kappa_\ell\,r_i^\top\Sigma_\ell r_i,
\qquad \kappa_\ell>0,
\label{eq:distortion-surrogate}
\end{equation}
where $\kappa_\ell$ is shared by the two coordinates, and cross-coordinate quantization-error terms are neglected.

\begin{proposition}[Guidance-weighted distortion]
Under this approximation, the expected squared branch-space distortion along
the CFG direction is proportional to
\vspace{-2mm}
\begin{equation}
D_{\mathrm{sur}}(R)
=
\sum_{i=1}^{2}
\underbrace{(r_i^\top g_s)^2}_{
  \substack{
    \text{\normalfont guidance}\\
    \text{\normalfont sensitivity}
  }
}
\quad
\underbrace{(r_i^\top \Sigma_\ell r_i)}_{
  \substack{
    \text{\normalfont projected}\\
    \text{\normalfont second moment}
  }
}.
\label{eq:gcbt-objective}
\end{equation}
\vspace{-4mm}
\end{proposition}

The objective therefore favors a branch basis that jointly accounts for how strongly each coding direction is represented in the activations and how strongly it contributes to the CFG. 

Because the branch space is two-dimensional, the rotation can be
parameterized by a single angle $\theta$. We measure all angles relative to
the native conditional axis and write
\begin{equation}
R(\theta)
=
\begin{bmatrix}
r_1(\theta)^\top\\
r_2(\theta)^\top
\end{bmatrix},
\qquad
r_1(\theta)=
\begin{bmatrix}
\cos\theta\\
\sin\theta
\end{bmatrix},
\quad
r_2(\theta)=
\begin{bmatrix}
-\sin\theta\\
\cos\theta
\end{bmatrix}.
\label{eq:rotation-param}
\end{equation}
Let $\psi_\ell$ denote the angle of the dominant eigenvector of
$\Sigma_\ell$, and let
$\phi_s=\operatorname{arctan2}(1-s,s)$ denote the angle of the CFG direction $g_s$.


\begin{theorem}[Closed-form optimal branch rotation]
\label{thm:gcbt}
Let $\lambda_{1,\ell}>\lambda_{2,\ell}$ be the eigenvalues of $\Sigma_\ell$. The global minimizers of Eq.~\ref{eq:gcbt-objective} over two-dimensional orthogonal rotations are
\begin{equation}
\theta_\ell^\star(s)
=
\frac{\phi_s+\psi_\ell}{2}
-
\frac{\pi}{4}
\qquad
\left(\operatorname{mod}\frac{\pi}{2}\right).
\label{eq:gcbt}
\end{equation}
If $\lambda_{1,\ell}=\lambda_{2,\ell}$, the surrogate is rotationally invariant.
\end{theorem}

We call $R(\theta_\ell^\star)$ the
\textbf{\emph{Guidance-Correlation Branch Transform} (GCBT).} A guidance-aligned basis depends only on $g_s$, while the principal-axis basis of $\Sigma_\ell$ depends only on activation geometry. GCBT combines both quantities in a single coding basis. Calibration requires only the three distinct entries of the $2\times2$ matrix $\Sigma_\ell$ per layer; no gradient optimization or angle search is required. The derivation and proof are given in Appendix~\ref{app:gcbt-proof} and~\ref{app:gcbt-theorem-proof}.

\subsection{Compatibility with Rotation-Based and CFG-Aware PTQ Backbones}
\label{sec:integration}

\paragraph{\textbf{\textit{Rotation-based Quantizer Compatibility.}}}
Some diffusion PTQ backbones apply a shared feature-space rotation or linear
transform before activation quantization. Let
$U_\ell\in\mathbb{R}^{d\times d}$ denote such a transform at layer $\ell$,
shared by the conditional and unconditional branches, and let
$R_\ell\in\mathbb{R}^{2\times2}$ denote the GCBT branch transform.
Because $U_\ell$ acts on the feature dimension whereas $R_\ell$ acts on the
CFG branch dimension, the two operations commute exactly.

\newpage

\begin{proposition}[Exact composition with shared feature-space rotations]
\label{prop:rotation-compat}
For branch activations
$X\in\mathbb{R}^{2\times N\times d}$,
a shared feature-space transform $U_\ell\in\mathbb{R}^{d\times d}$,
and a branch-space transform $R_\ell\in\mathbb{R}^{2\times2}$,
\begin{equation}
R_\ell(XU_\ell)=(R_\ell X)U_\ell .
\label{eq:rotation-commutation}
\end{equation}
Consequently, a host activation quantizer that operates on $XU_\ell$
can be combined with GCBT without changing the transformed activation seen by
the quantizer:
\begin{equation}
\mathcal{Q}_A\!\left(R_\ell(XU_\ell)\right)
=
\mathcal{Q}_A\!\left((R_\ell X)U_\ell\right).
\label{eq:rotation-host-eq}
\end{equation}
\end{proposition}

Thus, GCBT is algebraically compatible with host PTQ methods that already use
shared feature-space rotations: branch-space and feature-space transforms are
strictly composable rather than competing for the same degree of freedom.

\paragraph{\textbf{CFG-aware calibration hosts.}}
CFG-aware calibration and GCBT act on distinct quantization variables:
a guidance-aware host optimizes quantization parameters against guided-output
error while retaining the native $(c,u)$ coordinates, whereas GCBT changes
the branch basis in which quantization is performed. For a matched
branch-activation pair $x$ with $z=R_\ell x$,
\begin{equation}
g_s^\top x
= g_s^\top R_\ell^\top z
= (R_\ell g_s)^\top z .
\label{eq:cfg-readout-transformed}
\end{equation}
Thus, the CFG branch information is preserved exactly under the basis change, so guided-output calibration does not determine the branch coding basis.

\begin{wraptable}{r}{0.48\linewidth}
\vspace{-0.8em}
\centering
\footnotesize
\setlength{\tabcolsep}{3.5pt}
\caption{\textbf{GCBT after CFG-aware calibration.}
PixArt-$\Sigma$, W4A4, we implement~\citep{shafi2026closing} on SVDQuant given GAMP is closed-source. Bold: 95\%
CI excludes zero.}
\label{tab:cfg-aware-gcbt}
\vspace{-0.3em}
\begin{tabular}{llccc}
\toprule
Data & Method &
LPIPS $\downarrow$ &
SSIM $\uparrow$ &
PSNR $\uparrow$ \\
\midrule
\multirow{2}{*}{MJHQ}
& CFG-Cal. & 0.4775 & 0.5722 & 16.454 \\
& +GCBT & \textbf{0.4634} & \textbf{0.5829} & 16.547 \\
\midrule
\multirow{2}{*}{sDCI}
& CFG-Cal. & 0.5310 & 0.4768 & 15.279 \\
& +GCBT & \textbf{0.5179} & \textbf{0.4872} & 15.352 \\
\bottomrule
\end{tabular}
\vspace{-0.7em}
\end{wraptable}

We test this complementarity by applying GCBT to a fixed CFG-aware
W4A4 PixArt-$\Sigma$ host on MJHQ and sDCI, holding the calibrated host
parameters, quantization configuration, and generation seeds fixed.
Table~\ref{tab:cfg-aware-gcbt} shows consistent LPIPS and SSIM gains on
both datasets, while the smaller yet positive PSNR changes are not statistically conclusive. The gains therefore persist after guided-output calibration, supporting branch-basis selection as a complementary quantization variable. Full calibration details and paired confidence intervals are reported in
Appendix~\ref{app:cfg-aware-case-study}.

\FloatBarrier

\definecolor{oursblue}{RGB}{222,241,245}

\newcommand{\metricci}[3]{%
\begin{tabular}[c]{@{}c@{}}%
\textbf{#1}\\[-3pt]
{\fontsize{5.5}{6.5}\selectfont\color{black!65}$[#2,#3]$}%
\end{tabular}}

\newcommand{\metricciNB}[3]{%
\begin{tabular}[c]{@{}c@{}}%
#1\\[-3pt]
{\fontsize{5.5}{6.5}\selectfont\color{black!65}$[#2,#3]$}%
\end{tabular}}

\newcommand{\backendsetting}[2]{%
\begin{tabular}[c]{@{}c@{}}%
#1\\[-1pt]
#2%
\end{tabular}}

\newcommand{\modelsetting}[2]{%
\begin{tabular}[c]{@{}c@{}}%
#1\\[-1pt]
CFG #2%
\end{tabular}}

\newcommand{\pendingmetric}{\textit{TBD}}

\begin{table*}[ht]
\centering

\caption{
\textbf{Quantitative evaluation of image-generation fidelity on MJHQ and sDCI ($n=256$).} 
For each +GCBT row, the second line reports the pointwise paired 95\% confidence interval of improvement (Native minus GCBT for LPIPS; GCBT minus Native for SSIM and PSNR). Bold denotes gains whose paired interval lies entirely above zero. 
}
\label{tab:main_image}

\begin{threeparttable}
\resizebox{\linewidth}{!}{%
\small
\setlength{\tabcolsep}{4.0pt}
\renewcommand{\arraystretch}{1.20}

\begin{tabular}{ccccccccc}
\toprule
PTQ Hosts & Model & Method &
\multicolumn{3}{c}{MJHQ} &
\multicolumn{3}{c}{sDCI} \\
\cmidrule(lr){4-6}\cmidrule(lr){7-9}
&&&
LPIPS $\downarrow$ & SSIM $\uparrow$ & PSNR $\uparrow$ &
LPIPS $\downarrow$ & SSIM $\uparrow$ & PSNR $\uparrow$ \\
\midrule

\multirow{6}{*}{
  \backendsetting{SVDQuant}{W4A4}}
& \multirow{3}{*}{
  \modelsetting{PixArt-$\Sigma$}{4.5}}


& Native
& 0.327 & 0.651 & 17.472
& 0.355 & 0.574 & 16.450 \\

& & \cellcolor{oursblue}+ GCBT
& \cellcolor{oursblue}\metricci{0.296}{0.024}{0.038}
& \cellcolor{oursblue}\metricci{0.695}{0.039}{0.050}
& \cellcolor{oursblue}\metricci{17.917}{0.287}{0.607}
& \cellcolor{oursblue}\metricci{0.319}{0.029}{0.043}
& \cellcolor{oursblue}\metricci{0.625}{0.047}{0.057}
& \cellcolor{oursblue}\metricci{17.017}{0.418}{0.708} \\

\cmidrule(lr){2-9}

& \multirow{3}{*}{
  \modelsetting{SANA-1.6B}{4.5}}
& Native
& 0.230 & 0.702 & 17.283
& 0.234 & 0.651 & 16.565 \\

& & \cellcolor{oursblue}+ GCBT
& \cellcolor{oursblue}\metricci{0.210}{0.015}{0.026}
& \cellcolor{oursblue}\metricci{0.730}{0.023}{0.033}
& \cellcolor{oursblue}\metricci{17.825}{0.383}{0.703}
& \cellcolor{oursblue}\metricci{0.213}{0.016}{0.025}
& \cellcolor{oursblue}\metricci{0.685}{0.029}{0.038}
& \cellcolor{oursblue}\metricci{17.051}{0.364}{0.608} \\

\midrule

\multirow{6}{*}{
  \backendsetting{Q-DiT}{W4A8}}
& \multirow{3}{*}{
  \modelsetting{PixArt-$\Sigma$}{4.5}}
& Native
& 0.269 & 0.711 & 17.939
& 0.281 & 0.648 & 17.135 \\

& & \cellcolor{oursblue}+ GCBT
& \cellcolor{oursblue}\metricci{0.266}{0.001}{0.005}
& \cellcolor{oursblue}\metricci{0.715}{0.003}{0.006}
& \cellcolor{oursblue}\metricci{18.009}{0.035}{0.106}
& \cellcolor{oursblue}\metricci{0.278}{0.001}{0.005}
& \cellcolor{oursblue}\metricci{0.654}{0.004}{0.007}
& \cellcolor{oursblue}\metricci{17.192}{0.019}{0.094} \\

\cmidrule(lr){2-9}

& \multirow{3}{*}{
  \modelsetting{SANA-1.6B}{4.5}}
& Native
& 0.175 & 0.764 & 18.907
& 0.178 & 0.722 & 17.941 \\

& & \cellcolor{oursblue}+ GCBT
& \cellcolor{oursblue}\metricciNB{0.174}{-0.002}{0.004}
& \cellcolor{oursblue}\metricciNB{0.765}{-0.002}{0.003}
& \cellcolor{oursblue}\metricciNB{18.877}{-0.120}{0.059}
& \cellcolor{oursblue}\metricciNB{0.179}{-0.002}{0.002}
& \cellcolor{oursblue}\metricciNB{0.722}{-0.002}{0.002}
& \cellcolor{oursblue}\metricciNB{17.928}{-0.089}{0.057} \\

\midrule
\multicolumn{9}{c}{\textbf{\textit{Large-scale validation}}} \\
\midrule

\multirow{3}{*}{
  \backendsetting{SVDQuant}{W4A4}}
& \multirow{3}{*}{
  \modelsetting{SD3.5-Large}{4.5}}
& Native
& 0.394 & 0.674 & 16.242
& 0.386 & 0.653 & 15.898 \\

& & \cellcolor{oursblue}+ GCBT
& \cellcolor{oursblue}\metricci{0.378}{0.011}{0.021}
& \cellcolor{oursblue}\metricci{0.690}{0.013}{0.019}
& \cellcolor{oursblue}\metricci{16.385}{0.042}{0.243}
& \cellcolor{oursblue}\metricci{0.371}{0.010}{0.020}
& \cellcolor{oursblue}\metricci{0.667}{0.011}{0.018}
& \cellcolor{oursblue}\metricci{16.018}{0.031}{0.211} \\




\bottomrule
\end{tabular}%
}
\end{threeparttable}
\end{table*}

\definecolor{oursblue}{RGB}{222,241,245}
\providecommand{\videobackendsetting}[2]{%
  \begin{tabular}[c]{@{}c@{}}#1\\[-1pt]#2\end{tabular}}
\providecommand{\videomodelsetting}[2]{%
  \begin{tabular}[c]{@{}c@{}}#1\\[-1pt]CFG #2\end{tabular}}
\begin{table*}[ht]
\centering
\caption{
\textbf{Quantitative evaluation of video generation on VBench ($n=256$).} 
Bold indicates statistically significant improvements (paired 95\% CIs lie entirely in the favorable direction). 
Full paired confidence intervals are provided in Appendix Table~\ref{tab:video-ci}.
}
\label{tab:video-main}
\begin{threeparttable}
\resizebox{\linewidth}{!}{%
\small
\setlength{\tabcolsep}{4.0pt}
\renewcommand{\arraystretch}{1.20}
\begin{tabular}{ccccccccc}
\toprule
PTQ Host & Model & Method &
\multicolumn{3}{c}{Teacher Fidelity} & \multicolumn{3}{c}{VBench} \\
\cmidrule(lr){4-6}\cmidrule(lr){7-9}
&&& LPIPS $\downarrow$ & SSIM $\uparrow$ & PSNR $\uparrow$ &
S.C. $\uparrow$ & A.Q. $\uparrow$ & I.Q. $\uparrow$ \\
\midrule
\multirow{4}{*}{\videobackendsetting{OrbitQuant}{W4A4}} & \multirow{2}{*}{\videomodelsetting{Wan2.1-T2V-1.3B}{5}} & Native & 0.545 & 0.497 & 12.871 & 0.957 & 0.527 & 0.672 \\
 &  & \cellcolor{oursblue}+ GCBT & \cellcolor{oursblue}\textbf{0.504} & \cellcolor{oursblue}\textbf{0.518} & \cellcolor{oursblue}\textbf{12.965} & \cellcolor{oursblue}\textbf{0.975} & \cellcolor{oursblue}\textbf{0.584} & \cellcolor{oursblue}0.675 \\

\cmidrule(lr){2-9}
 & \multirow{2}{*}{\videomodelsetting{CogVideoX-2B}{6}} & Native & 0.427 & 0.639 & 17.474 & 0.927 & 0.494 & 0.593 \\
 &  & \cellcolor{oursblue}+ GCBT & \cellcolor{oursblue}\textbf{0.390} & \cellcolor{oursblue}\textbf{0.649} & 17.554 & \cellcolor{oursblue}\textbf{0.937} & \cellcolor{oursblue}\textbf{0.525} & \cellcolor{oursblue}\textbf{0.606} \\

\midrule
\multirow{2}{*}{\videobackendsetting{ViDiT-Q}{W4A8-MP}} & \multirow{2}{*}{\videomodelsetting{OpenSora v1.2}{7}} & Native & 0.544 & 0.540 & 13.871 & 0.935 & 0.455 & 0.490 \\
 &  & \cellcolor{oursblue}+ GCBT & \cellcolor{oursblue}\textbf{0.541} & \cellcolor{oursblue}\textbf{0.542} & \cellcolor{oursblue}13.891 & \cellcolor{oursblue}\textbf{0.938} & \cellcolor{oursblue}\textbf{0.459} & \cellcolor{oursblue}0.493 \\

\midrule
\multicolumn{9}{c}{\textit{Large-scale validation}} \\
\midrule
\multirow{2}{*}{\videobackendsetting{OrbitQuant port}{W4A4}} & \multirow{2}{*}{\videomodelsetting{Wan2.1-T2V-14B}{5}} & Native & 0.431 & 0.590 & 15.021 & 0.971 & 0.588 & 0.704 \\
 &  & \cellcolor{oursblue}+ GCBT & \cellcolor{oursblue}\textbf{0.413} & \cellcolor{oursblue}\textbf{0.601} & \cellcolor{oursblue}15.029 & \cellcolor{oursblue}\textbf{0.976} & \cellcolor{oursblue}\textbf{0.607} & \cellcolor{oursblue}0.703 \\

\bottomrule
\end{tabular}%
}
\end{threeparttable}
\end{table*}


\begin{figure}[htbp]
  \centering
  \includegraphics[width=\linewidth]{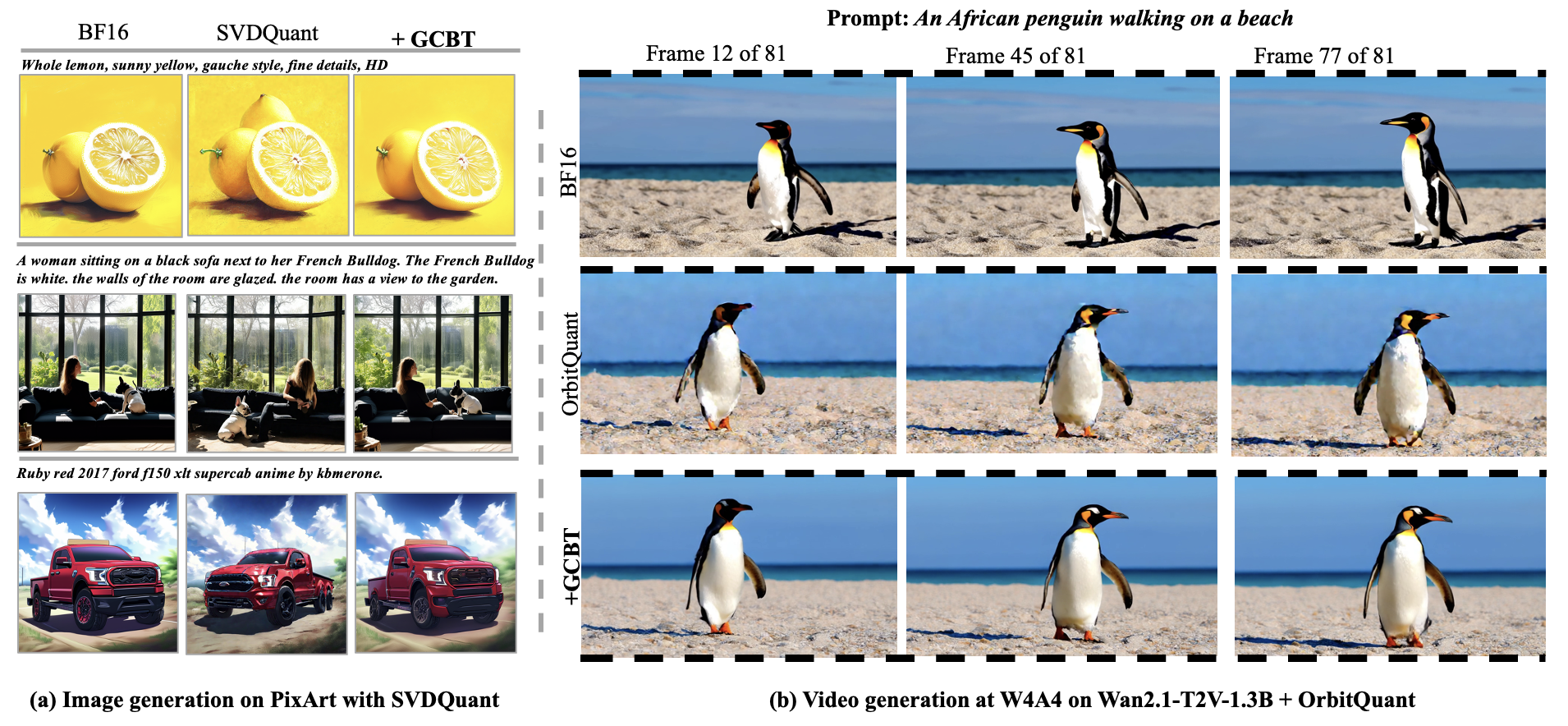}
  \caption{Qualitative comparison of GCBT on image and video generation.} 
  \label{fig:image-video-qualitative}
\end{figure}

\section{Experiments}
\label{sec:experiment}

\subsection{Experimental Settings}
\label{sec:exp-settings}


\paragraph{\textbf{Models and PTQ hosts.}}
We evaluate GCBT across image and video diffusion models with multiple
activation-quantization hosts. For image generation, the main evaluation
includes PixArt-$\Sigma$~\citep{chen2024pixart} and
SANA-1.6B~\citep{xie2024sana} with SVDQuant~\citep{li2026svdquant}
(W4A4) and Q-DiT~\citep{chen2024qdit} (W4A8), with SD3.5-Large~
\citep{esser2024scaling} included for large-scale validation.
All host precision settings strictly follow the default evaluation protocols established in their respective original works.
For video generation, we evaluate Wan2.1-T2V-1.3B and
Wan2.1-T2V-14B~\citep{wan2025}, CogVideoX-2B~\citep{yang2025cogvideox},
and OpenSora v1.2~\citep{zheng2024open} with OrbitQuant~
\citep{lee2026orbitquant} (W4A4) or ViDiT-Q~\citep{zhao2024viditq}
(W4A8-MP). GCBT uses 64 calibration samples disjoint from evaluation.
Full model-specific generation and calibration settings are provided in
Appendix~\ref{app:exp-details}.


\paragraph{Evaluation.}
We measure FP-reference fidelity using
LPIPS~\citep{zhang2018unreasonable},
SSIM~\citep{wang2004ssim}, and
PSNR~\citep{thu2008psnr}.
For video, we additionally report VBench subject consistency (S.C.),
aesthetic quality (A.Q.), and imaging quality (I.Q.)~
\citep{huang2024vbench}. Evaluation uses fixed paired prompt sets
($n=256$), and statistical significance is determined from paired
95\% confidence intervals; full effect sizes and intervals are reported
in the Appendix~\ref{app:exp-details}.

\subsection{Main Results}
\label{sec:main-results}

\paragraph{Image generation.}
Across the full image-generation matrix, GCBT yields statistically
significant gains in most paired metric comparisons, while no
comparison shows a statistically significant degradation. All
statistically resolved effects therefore favor GCBT. The significant
gains span both SVDQuant models, Q-DiT on PixArt-$\Sigma$, and the
larger SD3.5-Large model, consistently across MJHQ and sDCI. The remaining comparisons are statistically inconclusive, with no evidence of a systematic regression. Overall, these
results show that the benefits of GCBT transfer across architectures,
PTQ hosts, datasets, and model scales.

\paragraph{Video generation.}
The video experiments extend this pattern beyond image generation and,
more importantly, beyond teacher-referenced reconstruction alone.
Across Wan2.1, CogVideoX, and OpenSora, GCBT improves both
FP-referenced fidelity and VBench generation-quality metrics. LPIPS and
SSIM improve significantly for every evaluated video model, while
subject consistency and aesthetic quality also improve significantly
throughout. The effect persists on the larger Wan2.1-T2V-14B model,
showing that the branch-space benefit transfers across video
architectures, PTQ hosts, and model scales. These results indicate that
the fidelity gains from GCBT are accompanied by improvements in
generation-level quality, rather than being confined to closer
reconstruction of the full-precision reference.

\subsection{Ablations and System Analysis}
\label{sec:ablations}

We isolate the design choices underlying GCBT and examine its robustness
and systems cost. Unless stated otherwise, ablations use PixArt-$\Sigma$
with SVDQuant W4A4 on the fixed $n=256$ MJHQ subset at CFG $=4.5$, with
64 calibration samples disjoint from evaluation.

\paragraph{\textbf{Branch-basis components.}}
GCBT selects the coding basis jointly from the branch geometry
$\Sigma_\ell$ and the CFG readout direction $g_s$. We first compare the
native basis, a guidance-aligned basis using only $g_s$, and GCBT using
both quantities under the same quantization and generation settings.

\begin{wraptable}{r}{0.4\linewidth}
\vspace{-1.2em}
\centering
\scriptsize
\setlength{\tabcolsep}{2.7pt}
\caption{\textbf{Branch-basis ablation.}
PixArt-$\Sigma$, SVDQuant W4A4 on MJHQ. }
\label{tab:basis-ablation}
\vspace{-0.3em}
\resizebox{0.38\textwidth}{!}{%
\begin{tabular}{lccc}
\toprule
Basis &
LPIPS $\downarrow$ &
SSIM $\uparrow$ &
PSNR $\uparrow$ \\
\midrule
Native
& 0.327 & 0.651 & 17.472 \\

Guidance
& 0.315 & 0.678 & 17.707 \\

GCBT
& \textbf{0.296}
& \textbf{0.695}
& \textbf{17.917} \\
\bottomrule
\end{tabular}
}
\vspace{-1.7em}
\end{wraptable}

Table~\ref{tab:basis-ablation} shows a clear progression: aligning the
basis with the CFG readout already improves over native branch coding,
while incorporating the measured branch geometry yields a further gain
across all three fidelity metrics. This indicates that guidance
information alone does not account for the benefit of GCBT; the coding
basis benefits from coupling CFG sensitivity with the geometry of the
matched activations. Full statistics are reported in Appendix~\ref{app:basis-ablation}.

\begin{figure}[t]
    \centering
    \includegraphics[width=0.95\linewidth]{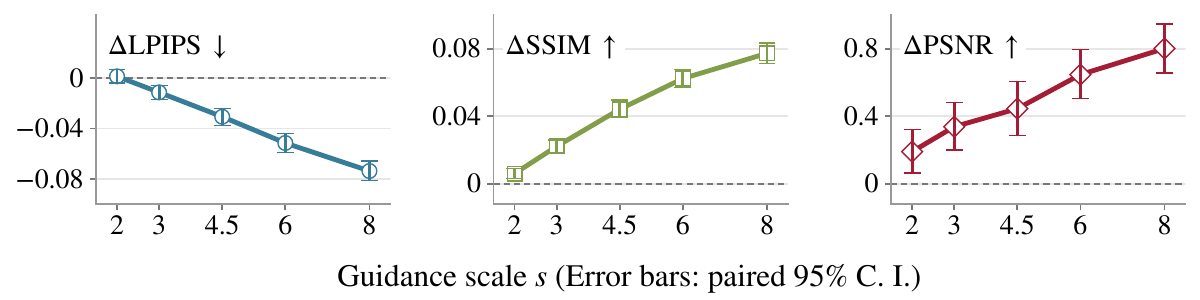}
    \caption{\textbf{Fidelity gains across CFG scales.} GCBT achieves monotonically larger gains over native PTQ across all metrics as guidance scale $s$ increases.}
    \label{fig:cfg-scale}
\end{figure}

\vspace{-0.2cm}
\paragraph{\textbf{Adaptation across guidance scales.}}
Because GCBT optimizes its basis using the CFG branch direction $g_s$, we evaluate its behavior across varying guidance scales while holding all other configurations fixed. As illustrated in Figure~\ref{fig:cfg-scale}, GCBT exhibits a clear scale-dependent pattern: its fidelity gains over the host quantizer grow monotonically as guidance becomes stronger. At weak guidance ($s=2$), the perceptual improvement is marginal; however, from $s=3$ onward, GCBT consistently strengthens performance across LPIPS, SSIM, and PSNR. This trend directly reflects the role of $g_s$ in our objective function: as stronger CFG alters the guidance readout orientation, adapting the branch coding basis becomes increasingly consequential for preserving activation fidelity. 

\begin{figure}[t]
    \centering
    \includegraphics[width=0.99\linewidth]{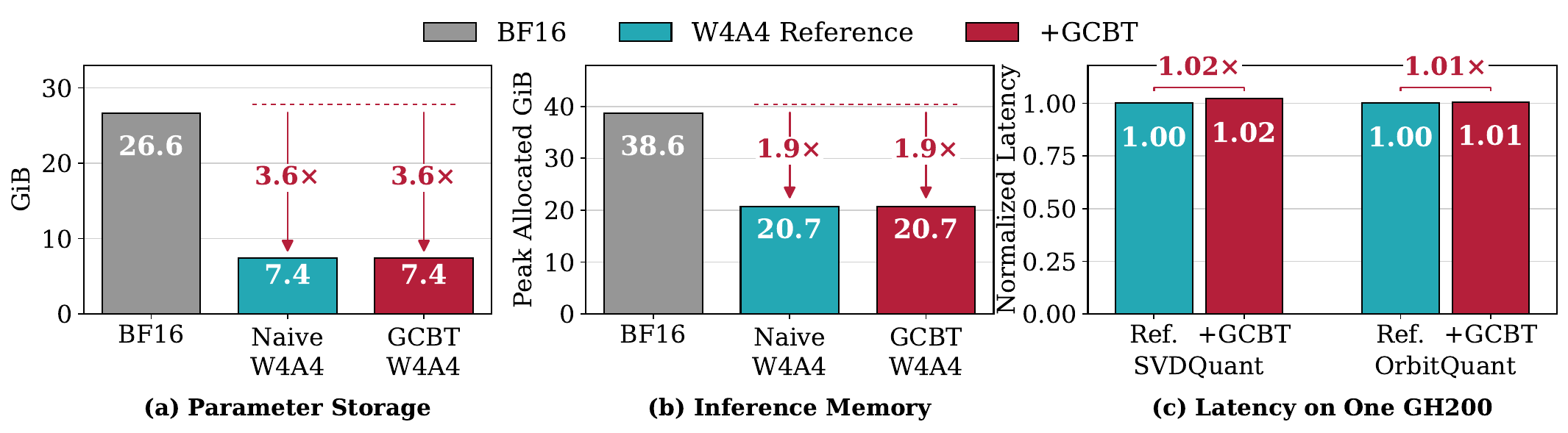}
    \caption{\textbf{System overhead comparison.} Host+GCBT matches native engines in end-to-end latency and peak GPU memory across image and video pipelines, imposing minimal practical overhead.}
    \label{fig:system-overhead}
\end{figure}

\vspace{-0.2cm}
\paragraph{\textbf{Runtime and memory overhead.}}
To evaluate GCBT's practical system cost, we measure end-to-end
generation latency and peak GPU memory footprint on representative
SVDQuant and OrbitQuant deployments, with SVDQuant evaluated using the
Nunchaku inference engine (offline calibration excluded).
As shown in Figure~\ref{fig:system-overhead}, adding GCBT leaves the
memory footprint unchanged and introduces only negligible runtime
overhead (1--2\%) relative to the corresponding host implementations.
Thus, the fidelity gains of GCBT do not require a materially more
expensive inference pipeline.

\begin{wraptable}{r}{0.38\linewidth}
\vspace{-1.25em}
\centering
\small 
\setlength{\tabcolsep}{5.0pt}
\caption{\textbf{Calibration stability.}
Layerwise medians.}
\label{tab:calibration-stability}
\vspace{-0.3em}
\resizebox{0.99\linewidth}{!}{%
\begin{tabular}{rccc}
\toprule
$n_{\mathrm{geom}}$ & Matched corr. & $\psi_\ell$ &
$\lambda_{1,\ell}/\lambda_{2,\ell}$ \\
\midrule
16  & 0.864 & $42.89^\circ$ & 13.71 \\
64  & 0.859 & $42.71^\circ$ & 13.27 \\
128 & 0.857 & $42.67^\circ$ & 13.03 \\
\bottomrule
\end{tabular}
}
\vspace{-0.8em}
\end{wraptable}


\vspace{-0.2cm}
\paragraph{\textbf{Calibration Sample Stability.}} GCBT depends only on the $2\times2$ branch second-moment matrix at each layer, so the required branch geometry can be estimated from a modest measurement pool. As shown in Table~\ref{tab:calibration-stability} (full results in Table~\ref{tab:motivation-calibration-sweep}), the measured geometry is already stable by $n_{\rm geom}=64$: increasing the pool to 128 changes the median principal-axis orientation by only $0.04^\circ$, while the matched-branch correlation remains approximately $0.86$ and the source remains strongly anisotropic. We therefore use 64 samples for GCBT calibration throughout. Full results are reported in Appendix~\ref{app:branch-calibration}.

\FloatBarrier

\section{Conclusion}

We identify CFG branch space as an overlooked coding dimension for diffusion activation quantization and introduce GCBT, a closed-form transform that jointly exploits matched-branch geometry and guidance sensitivity. Across many settings, GCBT consistently yields favorable fidelity and generation-quality improvements without modifying model parameters or the underlying quantization pipeline. These results establish branch-space coding as a lightweight and complementary design axis for diffusion PTQ, with negligible runtime and memory overhead.



\clearpage
\bibliographystyle{abbrvnat}
\bibliography{main}



\clearpage
\beginsupplement


\onecolumn
\appendix
{\hypersetup{linkcolor=black}
\tableofcontents
}

\begin{center}
    \LARGE {\bf Appendix}
\end{center}

\appendix

\section{Additional Details for GCBT}
\label{app:gcbt}

\subsection{Derivation of the Guidance-Weighted Distortion}
\label{app:gcbt-proof}

Fix a layer $\ell$ and suppress the layer superscript on local branch-space
quantities. Let $R=
\begin{bmatrix}
r_1^\top\\
r_2^\top
\end{bmatrix}
\in\mathbb{R}^{2\times2}$ be an orthogonal branch rotation, and let $z=Rx$ denote the transformed coordinates of a matched branch-activation pair $x$. Define the transformed-coordinate quantization error as $\delta
=
\widehat{z}-z
=
\begin{bmatrix}
\delta_1\\
\delta_2
\end{bmatrix},$ with $\delta_i=\widehat{z}_i-z_i.$

Following the distortion surrogate in Eq.~\ref{eq:distortion-surrogate}, we assume
\begin{equation}
\mathbb{E}[\delta_i^2]
=
\kappa_\ell\,r_i^\top\Sigma_\ell r_i,
\qquad
\mathbb{E}[\delta_1\delta_2]=0,
\qquad
\kappa_\ell>0,
\label{eq:app-distortion-surrogate}
\end{equation}
where the same $\kappa_\ell$ is used for both transformed coordinates.
No zero-mean assumption on the quantization error is required.

After inversion of the branch rotation, the local branch-space error is
$R^\top\delta$. Its component along the CFG readout direction $g_s$ is
\begin{equation}
g_s^\top R^\top\delta
=
\sum_{i=1}^{2}
(r_i^\top g_s)\delta_i.
\label{eq:app-guidance-error}
\end{equation}
Therefore,
\begin{align}
\mathbb{E}
\left[
\left(g_s^\top R^\top\delta\right)^2
\right]
&=
\sum_{i=1}^{2}
(r_i^\top g_s)^2
\mathbb{E}[\delta_i^2]
+
2(r_1^\top g_s)(r_2^\top g_s)
\mathbb{E}[\delta_1\delta_2]
\\
&=
\kappa_\ell
\sum_{i=1}^{2}
(r_i^\top g_s)^2
(r_i^\top\Sigma_\ell r_i).
\label{eq:app-guidance-distortion}
\end{align}
Because $\kappa_\ell>0$ is independent of the branch basis, minimizing this
expected squared distortion is equivalent to minimizing
$D_{\mathrm{sur}}(R)$ in Eq.~\ref{eq:gcbt-objective}.

\subsection{Proof of Theorem~\ref{thm:gcbt}}
\label{app:gcbt-theorem-proof}

Because the branch space is two-dimensional, any rotation can be
parameterized by a single angle $\theta$. Measuring angles relative to the
native conditional axis, write
\begin{equation}
R(\theta)
=
\begin{bmatrix}
r_1(\theta)^\top\\
r_2(\theta)^\top
\end{bmatrix},
\qquad
r_1(\theta)
=
\begin{bmatrix}
\cos\theta\\
\sin\theta
\end{bmatrix},
\qquad
r_2(\theta)
=
\begin{bmatrix}
-\sin\theta\\
\cos\theta
\end{bmatrix}.
\label{eq:app-rotation-param}
\end{equation}

Let
\begin{equation}
\bar{\lambda}_\ell
=
\frac{\lambda_{1,\ell}+\lambda_{2,\ell}}{2},
\qquad
\Delta_\ell
=
\frac{\lambda_{1,\ell}-\lambda_{2,\ell}}{2},
\label{eq:app-eigen-def}
\end{equation}
and let $\psi_\ell$ denote the angle of the dominant eigenvector of
$\Sigma_\ell$ and let $\phi_s$ denote the angle of $g_s$, both measured
relative to the native conditional axis. Then
\begin{align}
r_1^\top\Sigma_\ell r_1
&=
\bar{\lambda}_\ell
+
\Delta_\ell
\cos 2(\theta-\psi_\ell),
\\
r_2^\top\Sigma_\ell r_2
&=
\bar{\lambda}_\ell
-
\Delta_\ell
\cos 2(\theta-\psi_\ell).
\label{eq:app-projected-second-moment}
\end{align}
Similarly,
\begin{align}
r_1^\top g_s
&=
\|g_s\|_2
\cos(\phi_s-\theta),
\\
r_2^\top g_s
&=
\|g_s\|_2
\sin(\phi_s-\theta).
\label{eq:app-guidance-projection}
\end{align}

Substituting these expressions into
Eq.~\ref{eq:gcbt-objective} gives
\begin{align}
\frac{D_{\mathrm{sur}}(\theta)}
{\|g_s\|_2^2}
&\propto
\bar{\lambda}_\ell
+
\Delta_\ell
\cos 2(\phi_s-\theta)
\cos 2(\theta-\psi_\ell)
\\
&=
C_\ell
+
\frac{\Delta_\ell}{2}
\cos(4\theta-2\phi_s-2\psi_\ell),
\label{eq:app-objective-angle}
\end{align}
where
\[
C_\ell
=
\bar{\lambda}_\ell
+
\frac{\Delta_\ell}{2}
\cos 2(\phi_s-\psi_\ell)
\]
is independent of $\theta$.

For $\Delta_\ell>0$, the minimum is attained when
\begin{equation}
4\theta-2\phi_s-2\psi_\ell
=
-\pi
\qquad
(\operatorname{mod}2\pi),
\label{eq:app-opt-condition}
\end{equation}
which gives
\begin{equation}
\theta_\ell^\star(s)
=
\frac{\phi_s+\psi_\ell}{2}
-
\frac{\pi}{4}
\qquad
\left(
\operatorname{mod}\frac{\pi}{2}
\right).
\label{eq:app-opt-angle}
\end{equation}
This is Eq.~\ref{eq:gcbt}.
For $\Delta_\ell=0$, equivalently
$\lambda_{1,\ell}=\lambda_{2,\ell}$, the objective is independent of
$\theta$, and the surrogate is rotationally invariant.

\subsection{Commutation with Shared Linear Operators}
\label{app:commutation}

Let $\mathcal{L}(X)=XW$ be a linear operator shared by the two CFG branches.
Since $R$ acts along the branch dimension while $W$ acts along the feature
dimension,
\begin{equation}
\mathcal{L}(R^\top Z)
=
R^\top\mathcal{L}(Z).
\label{eq:app-commutation}
\end{equation}

Let $\widehat{Z}
=
\mathcal{Q}_A(RX)$ denote the quantized activation in transformed branch coordinates.
Equation~\ref{eq:app-commutation} allows the shared linear operator to act directly on $\widehat{Z}$,
\begin{equation}
\widehat{Z}
\xrightarrow{\;W\;}
Y'
\xrightarrow{\;R^\top\;}
Y,
\label{eq:app-lowbit-path}
\end{equation}
rather than first reconstructing the quantized activation in the native branch basis.

For an affine operator with a bias shared by the two CFG branches, the bias is added after the inverse branch transform. Thus, branch-space coding does not require reconstruction of the full native-basis activation before the shared quantized linear operation.

\section{Additional Experiments}
\label{sec:add_exp}

\subsection{Calibration-Size Stability}
\label{app:branch-calibration}

GCBT estimates the three distinct entries of the $2\times2$ second-moment matrix $\Sigma_\ell$ at each layer. To assess how much calibration data is required for these statistics to stabilize, we vary the calibration-set size on DiT-XL/2 with SVDQuant W4A4 and report layerwise median branch-space statistics. The mismatched correlation is computed by deliberately pairing conditional and unconditional activations from different samples, providing a reference for the matched-branch
structure.


\begin{table}[t]
\centering
\caption{\textbf{Measurement-pool-size stability of CFG branch geometry on DiT-XL/2.}
Values are medians across 112 linear sites under repository-native SVDQuant W4A4
with an FP rank-32 branch. All sizes use nested class/noise inputs and frozen weights.}
\label{tab:motivation-calibration-sweep}
\small
\setlength{\tabcolsep}{4.5pt}
\renewcommand{\arraystretch}{1.10}
\begin{tabular}{rcccc}
\toprule
$n_{\mathrm{geom}}$ & Matched corr. & Mismatched corr. & $\psi_\ell$ & $\lambda_{1,\ell}/\lambda_{2,\ell}$ \\
\midrule
16 & 0.8635 & 0.3166 & $42.89^\circ$ & 13.71 \\
32 & 0.8604 & 0.3243 & $42.71^\circ$ & 13.46 \\
\textbf{64} & \textbf{0.8587} & \textbf{0.3276} & $\mathbf{42.71^\circ}$ & \textbf{13.27} \\
128 & 0.8569 & 0.3231 & $42.67^\circ$ & 13.03 \\
\bottomrule
\end{tabular}
\end{table}


The estimated branch geometry is already stable at modest calibration sizes. In particular, beyond $n_{\mathrm{cal}}=64$, the principal-axis orientation changes negligibly, while matched-branch correlation remains close to $0.855$ and consistently exceeds the mismatched reference. The source also remains strongly anisotropic throughout the sweep. We therefore use 64 calibration samples in the main experiments.

\subsection{Paired Statistics for CFG-Aware Calibration}
\label{app:cfg-aware-case-study}

\paragraph{Setup.}
GAMP~\citep{shafi2026closing} has no public implementation, so we
implement a GAMP-inspired guided-output calibration on the official
SVDQuant W4A4 PixArt-$\Sigma$ host. This is not an official GAMP
reproduction. The W4 GPTQ residual weights, smoothing, and FP rank-32
branch are frozen. For each of the 168 activation quantizers
(dynamic group-64 A4), we fit one positive clipping multiplier in
$[0.1, 2.0]$ by minimizing the MSE between the quantized and
full-precision guided noise predictions, $\epsilon_u + s(\epsilon_c -
\epsilon_u)$ with $s = 4.5$. We use 1{,}024 paired states from 64
training prompts and 256 states from 16 validation prompts, 800 Adam
updates (learning rate $0.01$, batch size one), and keep the checkpoint
with the lowest validation error. The calibrated host is then frozen,
and GCBT angles are computed from 64 complete Native trajectories on
this host. Host-training, validation, GCBT-calibration, and evaluation
prompts are disjoint.

This study uses a separate execution setup from Table~\ref{tab:main_image}:
BF16 execution and a BF16 teacher, with initial noise drawn from a CPU
generator. Evaluation uses the same 256 MJHQ and 256 sDCI prompts and
seeds (1024$\times$1024, 20 DPM-Solver++ steps), but the different
dtype and noise generator yield different teacher images. Absolute
values are therefore not comparable across the two tables; the
evidence is the paired Host versus +GCBT comparison under an identical
host, prompts, seeds, and teacher.



\begin{table}[t]
\centering
\footnotesize
\setlength{\tabcolsep}{5.0pt}
\caption{\textbf{Paired effects of GCBT after CFG-aware calibration.}
Positive values favor GCBT. LPIPS reports relative reduction, while
SSIM and PSNR report absolute paired improvements. Brackets denote
paired 95\% confidence intervals.}
\label{tab:cfg-aware-gcbt-paired}
\begin{tabular}{lccc}
\toprule
Dataset &
LPIPS reduction $\uparrow$ &
$\Delta$SSIM $\uparrow$ &
$\Delta$PSNR (dB) $\uparrow$ \\
\midrule
MJHQ
& $2.97\%\;[2.17,\,3.78]\%$
& $+0.0107\;[0.0079,\,0.0135]$
& $+0.093\;[-0.005,\,0.190]$ \\
sDCI
& $2.47\%\;[1.71,\,3.23]\%$
& $+0.0104\;[0.0077,\,0.0132]$
& $+0.073\;[-0.017,\,0.169]$ \\
\bottomrule
\end{tabular}
\end{table}

As shown in Table~\ref{tab:cfg-aware-gcbt-paired}, the paired intervals
are entirely favorable for LPIPS and SSIM on both datasets, whereas the
PSNR intervals overlap zero. These statistics provide the paired
uncertainty estimates underlying the significance annotations in
Table~\ref{tab:cfg-aware-gcbt}.

\begin{table}[t]
\centering
\caption{
\textbf{Additional statistics for branch-basis ablation.}
Evaluated on PixArt-$\Sigma$ (SVDQuant W4A4, CFG $s=4.5$) on MJHQ ($n=256$).
Panel (a) reports absolute metric means with 95\% CIs.
Panel (b) reports paired improvements over the reference (positive values indicate improvement).
}
\label{tab:basis-ablation-stats}
\small
\setlength{\tabcolsep}{5.0pt}
\renewcommand{\arraystretch}{1.15}
\resizebox{\linewidth}{!}{%
\begin{tabular}{lccc}
\toprule
\multicolumn{4}{l}{
\textbf{(a) Absolute metrics: mean [95\% CI]}
} \\
\midrule
Basis &
LPIPS $\downarrow$ &
SSIM $\uparrow$ &
PSNR $\uparrow$ \\
\midrule
Native
& $0.327\,[0.315,\,0.338]$
& $0.651\,[0.638,\,0.664]$
& $17.472\,[17.150,\,17.805]$ \\

Guidance
& $0.315\,[0.303,\,0.327]$
& $0.678\,[0.664,\,0.691]$
& $17.707\,[17.378,\,18.044]$ \\

GCBT
& $\mathbf{0.296}\,[0.284,\,0.308]$
& $\mathbf{0.695}\,[0.682,\,0.708]$
& $\mathbf{17.917}\,[17.584,\,18.256]$ \\

\midrule
\multicolumn{4}{l}{
\textbf{(b) Paired improvements: mean [95\% CI]}
} \\
\midrule
Candidate vs.\ reference &
LPIPS reduction $\uparrow$ &
SSIM gain $\uparrow$ &
PSNR gain $\uparrow$ \\
\midrule
Guidance vs.\ Native
& $+0.012\,[0.004,\,0.020]$
& $+0.027\,[0.021,\,0.032]$
& $+0.236\,[0.069,\,0.406]$ \\

GCBT vs.\ Native
& $+0.031\,[0.024,\,0.038]$
& $+0.044\,[0.039,\,0.050]$
& $+0.445\,[0.287,\,0.607]$ \\

GCBT vs.\ Guidance
& $+0.019\,[0.014,\,0.024]$
& $+0.017\,[0.014,\,0.020]$
& $+0.209\,[0.104,\,0.307]$ \\
\bottomrule
\end{tabular}%
}
\end{table}

\subsection{Additional Statistics for Branch-Basis Ablation}
\label{app:basis-ablation}

Table~\ref{tab:basis-ablation-stats} provides the statistical details
underlying the branch-basis ablation in
Table~\ref{tab:basis-ablation}. All variants are evaluated on the same
256 MJHQ prompts with identical generation seeds, enabling paired
prompt-level comparisons. We report bootstrap 95\% confidence intervals
from 10,000 prompt-level resamples for both the absolute metric means and
the paired improvements between basis choices.

The paired analysis supports the stepwise pattern observed in the main
text. Guidance alignment significantly improves over Native in LPIPS,
SSIM, and PSNR, and GCBT provides a further statistically supported
improvement over the guidance-aligned basis on all three metrics.
Consequently, the full GCBT--Native comparison is also favorable
throughout. We base these comparisons on the paired intervals in
Table~\ref{tab:basis-ablation-stats}(b), rather than overlap between the
marginal confidence intervals of the absolute means in panel (a).

\subsection{Full Experimental Protocol \& Details}
\label{app:exp-details}



\paragraph{Image generation.}
PixArt-$\Sigma$ and SANA-1.6B are evaluated at
$1024\times1024$ resolution with 20 sampling steps and CFG $=4.5$. Both use the second-order DPM-Solver++ scheduler packaged with their respective checkpoints; PixArt-$\Sigma$ uses epsilon prediction, whereas SANA uses flow prediction. SD3.5-Large is evaluated at $1024\times1024$ resolution with 28 steps and CFG $=4.5$, using its checkpoint-provided FlowMatch Euler scheduler with shift $=3$. We use the official non-distilled checkpoints and fixed, empty negative prompts. Within each model, the full-precision reference, native quantized host, and GCBT use identical prompts, generation seeds, samplers, and generation settings.

\paragraph{Video generation.}
Wan2.1-T2V-1.3B and Wan2.1-T2V-14B generate $832\times480$ videos with 81 frames at 16 FPS, using 50 FlowUniPC steps and CFG $=5$; their flow shifts are 8 and 5, respectively. CogVideoX-2B generates $720\times480$ videos with 49 frames at 8 FPS, using 50 steps of its checkpoint-provided CogVideoX DDIM scheduler and CFG $=6$, with dynamic CFG disabled. OpenSora v1.2 generates $512\times512$ videos with 64 frames at 24 FPS, using 30 rectified-flow steps, timestep transformation, and CFG $=7$; its text preprocessing uses an AES score of 6.5. We use non-distilled checkpoints with active two-branch CFG throughout. For each model, the full-precision reference and both quantized variants share the prompt, generation seed, resolution, frame count, frame rate, and sampling configuration.

\definecolor{oursblue}{RGB}{222,241,245}
\providecommand{\videobackendsetting}[2]{%
  \begin{tabular}[c]{@{}c@{}}#1\\[-1pt]#2\end{tabular}}
\providecommand{\videomodelsetting}[2]{%
  \begin{tabular}[c]{@{}c@{}}#1\\[-1pt]CFG #2\end{tabular}}

\providecommand{\videometricci}[3]{%
  \begin{tabular}[c]{@{}c@{}}%
  #1\\[-3pt]%
  {\fontsize{5.5}{6.5}\selectfont\color{black!65}$[#2,\,#3]$}%
  \end{tabular}}
\providecommand{\videometricpending}[1]{%
  \begin{tabular}[c]{@{}c@{}}%
  #1\\[-3pt]%
  {\fontsize{6}{7}\selectfont\color{black!65}\textit{CI pending}}%
  \end{tabular}}
\par\medskip
\noindent\begin{minipage}{\linewidth}
\centering
\captionof{table}{\textbf{Paired confidence intervals for video-generation metrics on the fixed $n=256$ VBench subset.}
Each +GCBT cell reports the absolute mean, followed by the pointwise paired 95\% confidence interval of improvement:
Native minus GCBT for LPIPS, and GCBT minus Native for all other metrics.
Positive values favor GCBT. Bold marks favorable intervals excluding zero.
The intervals describe paired differences, not uncertainty intervals around the absolute means.}
\label{tab:video-ci}
\begin{threeparttable}
\resizebox{\linewidth}{!}{%
\small
\setlength{\tabcolsep}{4.0pt}
\renewcommand{\arraystretch}{1.20}
\begin{tabular}{ccccccccc}
\toprule
PTQ Host & Model & Method &
\multicolumn{3}{c}{Teacher Fidelity} & \multicolumn{3}{c}{VBench} \\
\cmidrule(lr){4-6}\cmidrule(lr){7-9}
&&& LPIPS $\downarrow$ & SSIM $\uparrow$ & PSNR $\uparrow$ &
S.C. $\uparrow$ & A.Q. $\uparrow$ & I.Q. $\uparrow$ \\
\midrule
\multirow{4}{*}{\videobackendsetting{OrbitQuant}{W4A4}} & \multirow{2}{*}{\videomodelsetting{Wan2.1-T2V-1.3B}{5}} & Native & 0.545 & 0.497 & 12.871 & 0.957 & 0.527 & 0.672 \\
 &  & \cellcolor{oursblue}+ GCBT & \cellcolor{oursblue}\videometricci{\textbf{0.504}}{0.036070}{0.046001} & \cellcolor{oursblue}\videometricci{\textbf{0.518}}{0.017282}{0.025529} & \cellcolor{oursblue}\videometricci{\textbf{12.965}}{0.017197}{0.179532} & \cellcolor{oursblue}\videometricci{\textbf{0.975}}{0.017177}{0.020618} & \cellcolor{oursblue}\videometricci{\textbf{0.584}}{0.051873}{0.063167} & \cellcolor{oursblue}\videometricci{0.675}{-0.002505}{0.007421} \\

\cmidrule(lr){2-9}
 & \multirow{2}{*}{\videomodelsetting{CogVideoX-2B}{6}} & Native & 0.427 & 0.639 & 17.474 & 0.927 & 0.494 & 0.593 \\
 &  & \cellcolor{oursblue}+ GCBT & \cellcolor{oursblue}\videometricci{\textbf{0.390}}{0.030245}{0.044470} & \cellcolor{oursblue}\videometricci{\textbf{0.649}}{0.004370}{0.015454} & \cellcolor{oursblue}\videometricci{17.554}{-0.093723}{0.251918} & \cellcolor{oursblue}\videometricci{\textbf{0.937}}{0.007323}{0.011380} & \cellcolor{oursblue}\videometricci{\textbf{0.525}}{0.026006}{0.036021} & \cellcolor{oursblue}\videometricci{\textbf{0.606}}{0.005985}{0.018786} \\

\midrule
\multirow{2}{*}{\videobackendsetting{ViDiT-Q}{W4A8-MP}} & \multirow{2}{*}{\videomodelsetting{OpenSora v1.2}{7}} & Native & 0.544 & 0.540 & 13.871 & 0.935 & 0.455 & 0.490 \\
 &  & \cellcolor{oursblue}+ GCBT & \cellcolor{oursblue}\videometricci{\textbf{0.541}}{0.001084}{0.003664} & \cellcolor{oursblue}\videometricci{\textbf{0.542}}{0.001102}{0.003845} & \cellcolor{oursblue}\videometricci{13.891}{-0.014214}{0.059996} & \cellcolor{oursblue}\videometricci{\textbf{0.938}}{0.000561}{0.004034} & \cellcolor{oursblue}\videometricci{\textbf{0.459}}{0.000861}{0.007971} & \cellcolor{oursblue}\videometricci{0.493}{-0.000222}{0.006053} \\

\midrule
\multicolumn{9}{c}{\textit{Large-scale validation}} \\
\midrule
\multirow{2}{*}{\videobackendsetting{OrbitQuant port}{W4A4}} & \multirow{2}{*}{\videomodelsetting{Wan2.1-T2V-14B}{5}} & Native & 0.431 & 0.590 & 15.021 & 0.971 & 0.588 & 0.704 \\
 &  & \cellcolor{oursblue}+ GCBT & \cellcolor{oursblue}\videometricci{\textbf{0.413}}{0.013673}{0.022822} & \cellcolor{oursblue}\videometricci{\textbf{0.601}}{0.007352}{0.015371} & \cellcolor{oursblue}\videometricci{15.029}{-0.088516}{0.103788} & \cellcolor{oursblue}\videometricci{\textbf{0.976}}{0.003256}{0.006483} & \cellcolor{oursblue}\videometricci{\textbf{0.607}}{0.014723}{0.023360} & \cellcolor{oursblue}\videometricci{0.703}{-0.003987}{0.002373} \\

\bottomrule
\end{tabular}%
}
\end{threeparttable}
\end{minipage}
\par\medskip

\paragraph{Paired evaluation protocol.}
For video, we use a fixed $n=256$ subset of
VBench~\citep{huang2024vbench}, stratified by category. Prompt subsets
and generation seeds are fixed once and reused for the full-precision
reference, Host PTQ, and Host+GCBT. Within each paired comparison, the
checkpoint, host configuration, bitwidth, CFG scale, sampling steps,
resolution, and generation seed are identical; only GCBT is changed.
Framewise fidelity metrics are averaged within each clip, with prompts
treated as the paired statistical units.

\begin{figure}[t]
    \centering
    \includegraphics[width=0.99\linewidth]{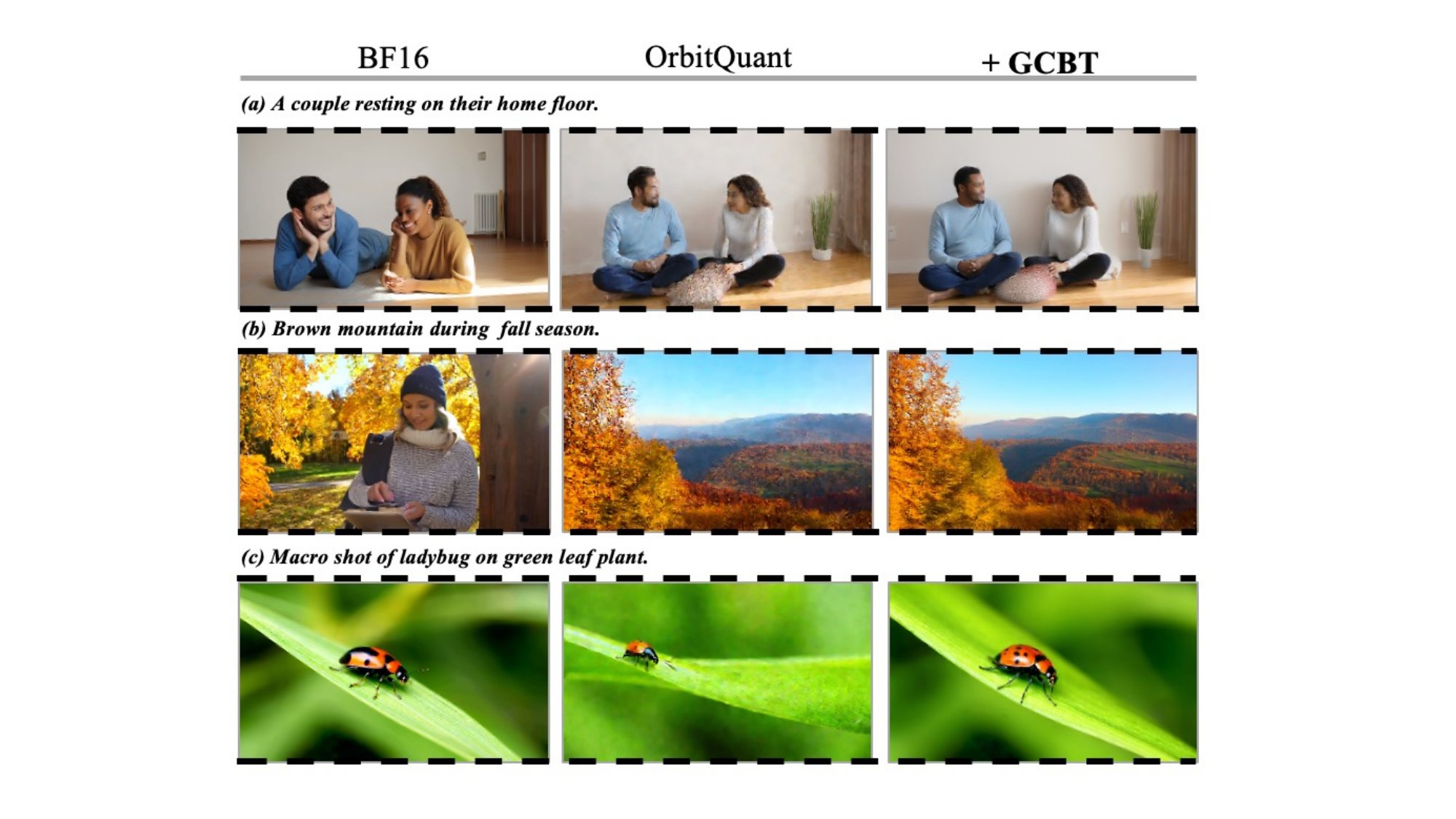}
    \captionof{figure}{\textbf{Additional qualitative examples for the appendix.}
    We show full-precision, native quantized, and GCBT outputs under the same prompts and generation settings.}
    \label{fig:appendix-b5}
\end{figure}

\paragraph{On the absolute LPIPS values for Wan.}
The relatively high BF16-referenced LPIPS on Wan is not specific to GCBT: it is already present in the OrbitQuant baseline. In video diffusion, quantization can shift the sampling trajectory enough to produce a different yet still prompt-consistent realization, making frame-aligned distance to a single BF16 sample large even when visual quality remains good. Figure~\ref{fig:appendix-b5} illustrates this behavior. Accordingly, we interpret LPIPS here as BF16-reference fidelity rather than standalone generation quality, and report VBench metrics alongside it.

\begin{figure}[t]
    \centering
    \includegraphics[width=0.99\linewidth]{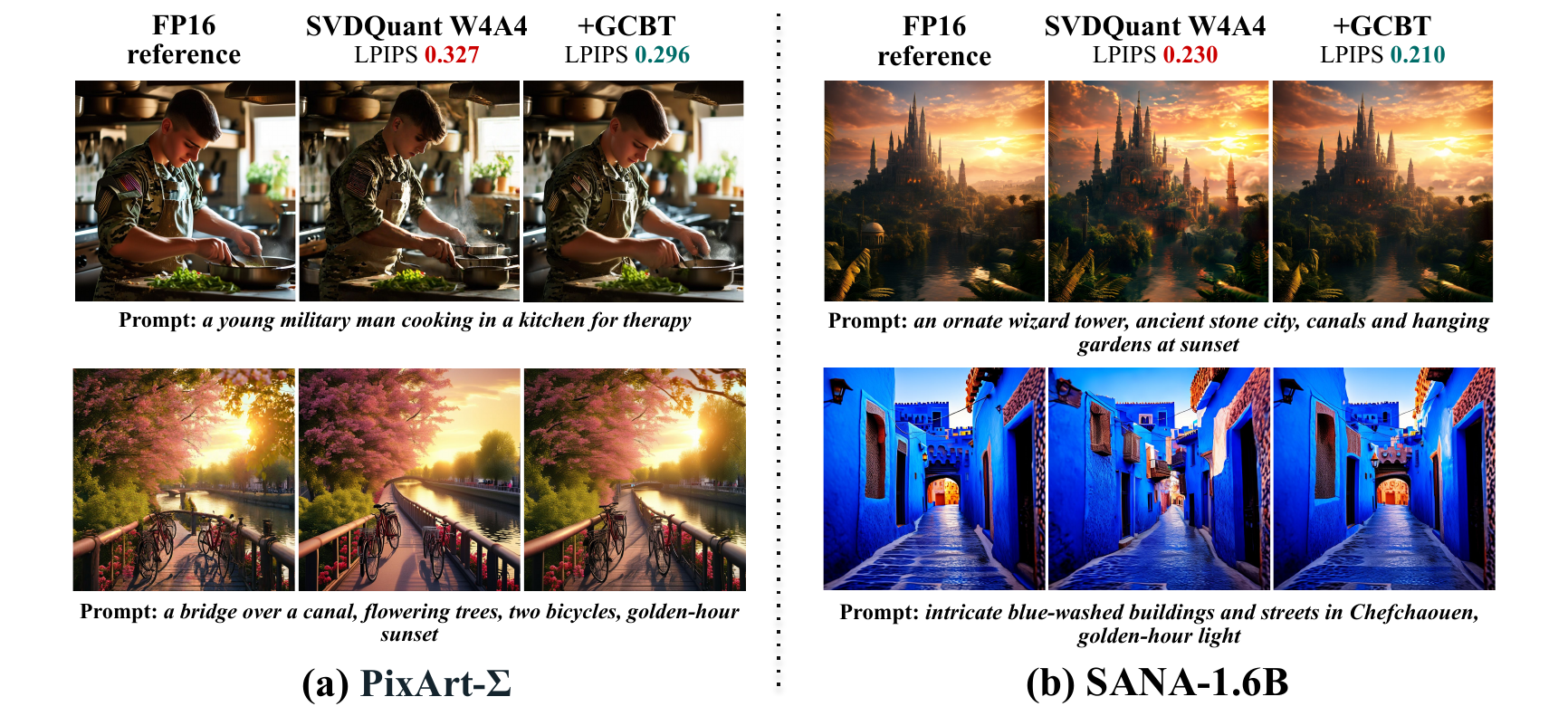}
    \caption{\textbf{More Examples of GCBT Generation}}
    \label{fig:more_examples}
\end{figure}

\section{Limitations}
\label{sec:limitations}

GCBT has several scope limitations. First, its closed-form optimality is derived
under the equal-rate quantization-noise surrogate in Sec.~3.3, which assumes
that transformed-coordinate quantization error scales with projected second
moments and neglects cross-coordinate error terms. The resulting rotation is
therefore optimal for this surrogate, rather than being guaranteed to minimize
the exact distortion of every host quantizer or end-to-end generation error.

Second, GCBT requires an explicit matched conditional--unconditional branch
pair and is therefore directly applicable to inference pipelines with active
two-branch CFG. We do not study distilled or single-pass guidance schemes, nor
timestep-varying or dynamic CFG schedules.

Finally, our CFG-aware comparison uses a GAMP-inspired calibration that we
implemented ourselves because no public implementation of GAMP is available.
This experiment should therefore be interpreted as evidence that GCBT remains
complementary to guided-output calibration, rather than as an official
reproduction of GAMP; results with the original implementation may differ.

\end{document}